\documentclass[letterpaper, 10 pt, conference]{ieeeconf}  

\IEEEoverridecommandlockouts                              

\usepackage{tikz}
\usepackage{pgfplots}
\usepackage{graphicx} 
\usepackage{setspace}
\usepackage{xcolor}
\usepackage{amsmath}
\usepackage{amsfonts}
\usepackage{bbm}
\usepackage{mathtools} 
\RequirePackage{algorithm}
\RequirePackage{algpseudocode} 
\usepackage{caption}
\usepackage{subcaption}
\usepackage{comment}
\usepackage{listings}
\usepackage{dsfont}
\usepackage{amsfonts}
\usepackage{booktabs}
\usepackage{optidef}
\usepackage{hyperref}
\usepackage{url}
\usepackage{bm}
\usepackage{tikz}
\usetikzlibrary{shapes, arrows.meta, positioning, calc}
\usepackage{comment}

\newcommand{\norm}[1]{\left\lVert#1\right\rVert}
\newcommand{\R}[1]{\mathrm{#1}}

\newcommand{\domain}[0]{\mathcal{D}}

\usepackage{dsfont}
\newcommand{\indicator}[1]{\mathds{1}_{\left\{#1\right\}}}

\newcommand{\B}[1]{\if#1\relax\bm{#1}\else\mathbf{#1}\fi} 
\newcommand{\BB}[1]{\mathbb{#1}}

\definecolor{myorange}{RGB}{205, 102, 57}
\pgfplotsset{compat=1.18}

\usepackage{csquotes}

\title{\LARGE \bf
Hierarchical Policy-Gradient Reinforcement Learning for Multi-Agent Shepherding Control of Non-Cohesive Targets
}

\author{Stefano Covone\textsuperscript{1}, Italo Napolitano\textsuperscript{1}, Francesco De Lellis\textsuperscript{2}, Mario di Bernardo\textsuperscript{1,2,*} 
\thanks{This work was developed with the economic support of MIUR (Italian Ministry of University and Research) performing the activities of the project PRIN 2022 “Machine-learning based control of complex multi-agent systems for search and rescue operations in natural disasters (MENTOR).}
\thanks{\textsuperscript{1}Scuola Superiore Meridionale, Naples, Italy}%
\thanks{\textsuperscript{2}Department of Electrical Engineering and
Information Technology, University of Naples Federico II, Naples, Italy}%
\thanks{\textsuperscript{*}Corresponding author: mario.dibernardo@unina.it}
\thanks{The authors wish to thank Andrea Lama from Scuola Superiore Meridionale for the useful discussions and support during the initial phase of this work.}
}

\begin{document}

\maketitle
\thispagestyle{empty}
\pagestyle{empty}


\begin{abstract}
We propose a decentralized reinforcement learning solution for multi-agent shepherding of non-cohesive targets using policy-gradient methods. Our architecture integrates target-selection with target-driving through Proximal Policy Optimization, overcoming discrete-action constraints of previous Deep Q-Network approaches and enabling smoother agent trajectories. This model-free framework effectively solves the shepherding problem without prior dynamics knowledge. Experiments demonstrate our method's effectiveness and scalability with increased target numbers and limited sensing capabilities.
\end{abstract}

\section{Introduction}
\label{sec:introduction}
The shepherding problem in robotics exemplifies the problem of harnessing complex systems for control \cite{lamaShepherdingControlHerdability2024, lamaInterpretableContinuumFramework2025}. It generally involves a group of actively controlled agents, termed \textit{herders}, strategically influencing a group of passive agents, referred to as \textit{targets}. From a control-theoretic perspective, this represents a paradigmatic example of indirect control in complex systems \cite{licitraSingleAgentIndirectherding2018}, where an active system manipulates a passive one to guide its collective behavior toward a desired state.

This problem has wide-ranging applications: from wildlife management using UAVs to prevent animals from entering dangerous areas \cite{paranjapeRoboticHerdingFlock2018}, to security tasks in search and rescue operations \cite{liuPlanningAssistedContextSensitiveAutonomous2023} and protection of strategic areas \cite{chipadeAerialSwarmDefense2021}.

State-of-the-art shepherding solutions face several challenges. Model-based approaches \cite{koAsymptoticBehaviorControl2020} create computationally intensive optimization problems that scale poorly, while heuristic methods \cite{strombomSolvingShepherdingProblem2014, lienShepherdingBehaviors2004a} typically use predefined two-phase strategies where herders first collect targets and then drive them toward a desired region.

In contrast, learning-based approaches have emerged as powerful alternatives for shepherding, enabling agents to develop effective strategies without predefined rules or prior knowledge of target dynamics. For example, in \cite{zhi2021learning}, herders learn to steer flocks in complex structured environments via Deep Q-Networks. Deep policy-gradient Reinforcement Learning (RL) has further advanced these capabilities. In \cite{nguyen2020continuousdeephierarchicalreinforcement}, separate driving and collecting policies are learnt using threshold-based switching, while curriculum learning is employed in \cite{husseinAutonomousSwarmShepherding2022}  for driving and mode-selection. These methods, however, are limited to single-herder scenarios.

For multi-herder coordination, Multi-Agent Reinforcement Learning (MARL) offers promising solutions \cite{guptaCooperativeMultiagentControl2017}.  Deep Deterministic Policy Gradient (DDPG) is applied in \cite{WangMultiDrone2024} for centralized learning of driving and collecting behaviors, while  Proximal Policy Optimization (PPO) with convolutional networks is used in \cite{hasanFlockNavigation2023} to develop scalable, decentralized policies.

Despite their impressive performance, these approaches rely on the simplifying assumption that targets behave as a cohesive flock \cite{strombomSolvingShepherdingProblem2014} -- a hypothesis that proves unrealistic in various domains including emergency evacuation scenarios, wildlife management of non-herding species, and heterogeneous robotic swarms \cite{lamaShepherdingControlHerdability2024}. Without cohesion, as noted in \cite{koAsymptoticBehaviorControl2020}, the problem complexity increases significantly as herders must control each target individually rather than influencing the group's center of mass.

To address this scenario, herder behavior is divided in \cite{aulettaHerdingStochasticAutonomous2022} into two distinct tasks: \textit{target-selection}, where herders decide what targets to chase, and \textit{driving}, where herders push the selected targets towards the goal region. 
Another approach \cite{delellisApplicationControlTutored2021} uses Q-learning with heuristic rules, while \cite{napolitano2024emergentcooperativestrategiesmultiagent} employs DQN but suffers from discrete action limitations.

In this work, we leverage Proximal Policy Optimization (PPO), a state-of-the-art deep policy-gradient algorithm \cite{schulmanProximalPolicyOptimization2017}, aiming to: (i) operate within a fully learning-based framework, overcoming the reliance on heuristic components found in hybrid approaches, (ii) require no prior knowledge of target dynamics, and (iii) achieve state-of-the-art performance by utilizing a continuous action space, rather than a discrete one as in DQN.

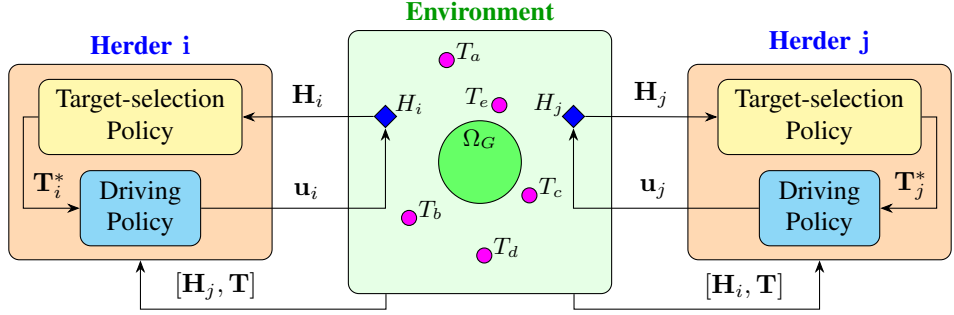
\begin{figure*}[ht]
    \centering
    \vspace{0.1cm}
        \begin{tikzpicture}[node distance=1cm, auto, >=Stealth]

        \definecolor{h_blue}{RGB}{0, 0, 255}
        \definecolor{t_magenta}{RGB}{255, 0, 255}
        \definecolor{g_green}{RGB}{102, 255, 102}

        \node[rectangle, draw, rounded corners, align=center, fill=green!10, minimum width=3.5cm,
              minimum height=3.5cm, label={[green!60!black, above, yshift=-0cm]\textbf{Environment}},] (env) {};

        \node[rectangle, draw, rounded corners, align=center, fill=orange!30, minimum width=3.5cm, minimum height=2.6cm, left=1cm of env, label={[h_blue, above]\textbf{Herder $\B{i}$}},] (agent_i) {};
        
        \node[rectangle, draw, rounded corners, fill=yellow!40,
              align=center, minimum width=2.7cm, minimum height=1cm] at ($(agent_i) + (0, 0.6)$) (selection_i) 
              {Target-selection\\Policy};

        \node[rectangle, draw, rounded corners, fill=cyan!40,
              align=center, minimum width=1.6cm, minimum height=1cm,
              below=0.2cm of selection_i] (driving_i) {Driving\\Policy};

        \node[rectangle, draw, rounded corners, align=center, fill=orange!30, minimum width=3.5cm, minimum height=2.6cm, right=1cm of env, label={[h_blue, above]\textbf{Herder $\B{j}$}},] (agent_j) {};
        
        \node[rectangle, draw, rounded corners, fill=yellow!40,
              align=center, minimum width=2.7cm, minimum height=1cm] at ($(agent_j) + (0, 0.6)$) (selection_j) 
              {Target-selection\\Policy};

        \node[rectangle, draw, rounded corners, fill=cyan!40,
              align=center, minimum width=1.6cm, minimum height=1cm,
              below=0.2cm of selection_j] (driving_j) {Driving\\Policy};

  \node[circle, draw, fill=g_green, minimum size=1.1cm, inner sep=0pt, label={[below, font=\small]$\Omega_G$}] 
    at (env.center) {};

  \node[diamond, draw, fill=h_blue, minimum size=0.3cm, inner sep=0pt,
      label={[right, font=\small]$H_i$}]
    at ([shift={(0.5,-1.15)}]env.north west) (herder_i) {};
  \node[diamond, draw, fill=h_blue, minimum size=0.3cm, inner sep=0pt,
      label={[left, font=\small]$H_j$}]
    at ([shift={(3,-1.15)}]env.north west) (herder_j) {};
  
  \node[circle, draw, fill=t_magenta, minimum size=0.2cm, inner sep=0pt,
      label={[right, font=\small]$T_a$}]
    at ([shift={(-2.2,-0.4)}]env.north east) {};
  \node[circle, draw, fill=t_magenta, minimum size=0.2cm, inner sep=0pt,
      label={[right, font=\small]$T_b$}]
    at ([shift={(-2.7,-2.5)}]env.north east) {};
    \node[circle, draw, fill=t_magenta, minimum size=0.2cm, inner sep=0pt,
      label={[right, font=\small]$T_c$}]
    at ([shift={(-1.1,-2.2)}]env.north east) {};
    \node[circle, draw, fill=t_magenta, minimum size=0.2cm, inner sep=0pt,
      label={[right, font=\small]$T_d$}]
    at ([shift={(-1.7,-3)}]env.north east) {};
    \node[circle, draw, fill=t_magenta, minimum size=0.2cm, inner sep=0pt,
      label={[left, font=\small]$T_e$}]
    at ([shift={(-1.5,-1)}]env.north east) {};

        \draw[->] (selection_i.west) -| ($(selection_i.west) - (0.2,0)$)  |- node[right, yshift=0.3cm] {$\B{T}_i^*$} (driving_i.west);

        \draw[->] (driving_i.east)  node[above, xshift=1.4cm] {$\B{u}_i$} -| (herder_i.south);

        \draw[->] ($(env.south) - (1.25,0)$)  -| ++(0,-0.2) -| node[above, xshift=1cm] {$[\B{H}_j, \B{T}]$} (agent_i.south);

        \draw[->] (herder_i.west) -- node[above, xshift=0cm] {$\B{H}_i$} (selection_i.east);

        \draw[->] (selection_j.east) -| ($(selection_j.east) + (0.2,0)$)  |- node[left, yshift=0.3cm] {$\B{T}_j^*$} (driving_j.east);

        \draw[->] (driving_j.west)  node[above, xshift=-1.4cm] {$\B{u}_j$} -| (herder_j.south);

        \draw[->] ($(env.south) + (1.25,0)$)  -| ++(0,-0.2) -| node[above, xshift=-1cm] {$[\B{H}_i, \B{T}]$} (agent_j.south);

        \draw[->] (herder_j.east) -- node[above] {$\B{H}_j$} (selection_j.west);

    \end{tikzpicture}

    \vspace{0.3cm}
    
    \caption{Two-layer feedback control scheme: each herder $\mathbf{H}_{i,j}$ detects the other agents' positions and  selects the target $\mathbf{T}_{i,j}^*$ to control via the \textit{target-selection} policy, which is then driven according to the \textit{driving} policy, that outputs the velocity $\B{u}$ of the corresponding herder.
    }
    \label{fig:control_architecture}
\end{figure*}

\section{Problem statement}
\label{sec: problem_statement}
Consider a square domain \(\domain = \left[-R,R\right]^2 \subset \mathbb{R}^2\), where \(N\) herders must guide \(M\) targets to a circular goal region \(\Omega_\R{G} \in \domain\), centered at the origin, with radius \(\rho_\R{G} < R\). Initially, both the herders and the targets are uniformly randomly distributed within a circle \(\Omega_\R{0} \in \domain\) of radius \(R\), centered at the origin.

Let \(\mathbf{H} \in \mathbb{R}^{2N}\) denote the vector of herder positions, defined as \(\mathbf{H} = [\mathbf{H}_1, \ldots, \mathbf{H}_N]\), where \(\mathbf{H}_i \in \mathbb{R}^2\) represents the Cartesian coordinates of the \(i\)-th herder. Similarly, let \(\mathbf{T} \in \mathbb{R}^{2M}\) denote the vector of target positions, defined as \(\mathbf{T} = [\mathbf{T}_1, \ldots, \mathbf{T}_M]\), where \(\mathbf{T}_a \in \mathbb{R}^2\) represents the Cartesian coordinates of the \(a\)-th target.

Building on the minimal shepherding model presented in \cite{lamaShepherdingControlHerdability2024}, both populations are modeled with first-order differential equations under the assumption of a negligible acceleration phase duration \cite{albi2016}, providing sufficient dynamic fidelity while maintaining computational efficiency for the learning framework. 
Unlike most existing studies, this model does not assume cohesive collective behavior among the targets; instead, their dynamics are governed by the overdamped Langevin equation:
\begin{equation}
\label{eqn:target_dyn}
    \dot {\mathbf{T}}_a = \sqrt{2D} \B{\eta} + k^\R{T} \sum_{i=1}^N \mathbf{\Theta} (\lambda, \mathbf{T}_a - \mathbf{H}_i)    
\end{equation}
where $\B{\eta} \in \BB{R}^2$ represents standard Gaussian noise and $\mathbf{\Theta}(\lambda, \mathbf{x})$ defines an inter-population interaction of the form
\begin{equation}
    \mathbf{\Theta}(\lambda, \mathbf{x}) = \begin{cases}
        (\lambda - |\mathbf{x}|)\hat{\mathbf{x}} & |\mathbf{x}| \leq \lambda \\
        0 & |\mathbf{x}| > \lambda
    \end{cases},
\end{equation}
representing a harmonic repulsion exerted by all nearby herders on the $a$-th target, which is typically adopted in Physics as the simplest type of repulsive force among particles \cite{dolayPhaseSeparation2018}. Specifically, herders repel targets within a characteristic distance $\lambda$, with intensity regulated by $k^\R{T}$ \cite{lamaShepherdingControlHerdability2024}. 
The maximum escape speed of a target under repulsion from a nearby herder is given by $v_\R{T} = k^\R{T} \lambda$, assuming $k^\R{T} \lambda^2 \gg D$ to ensure that the repulsive force dominates the targets' Brownian motion.

The herders are modeled as single integrators following the control law -- e.g. mobile robots with a feedback linearized dynamics -- as commonly used in the robotics literature for this problem \cite{piersonControllingNoncooperativeHerds2018}:
\begin{equation}
    \label{eqn: herder_dyn}
    \dot {\mathbf{H}}_i = \mathbf{u}_i
\end{equation}
where $\mathbf{u}_i \in \mathbb{R}^2$ represents the control input for the $i$-th herder, with a maximum velocity $v_\R{H} > k^\R{T} \lambda$ to prevent the formation of stable chasing patterns \cite{lamaShepherdingControlHerdability2024}. 
The model is simulated using the Euler–Maruyama integration method. The nominal parameters, provided in the Appendix, are taken from \cite{lamaShepherdingControlHerdability2024}, except for the diffusion coefficient $D$, which is set to a smaller value.

The goal is to design a control law $\mathbf{u}_i$ for $i=1, \dots, N$ such that every target is driven and contained within the designated goal region $\Omega_\R{G}$ in finite time. A buffered region $\hat{\Omega}_G$ with radius $(1 + \varepsilon)\rho_G$ is introduced to accommodate for stochastic effects. 
We consider a decentralized, communication-free setting where herders rely solely on their observations, and can sense agents in the entire domain but do not exchange internal information, i.e., their decisions, as considered in \cite{li2023communication}.

\section{Control architecture}

To manage the complexity of the task, we adopt a two-layer hierarchical approach, as also done in \cite{napolitano2024emergentcooperativestrategiesmultiagent}. As illustrated in Fig. \ref{fig:control_architecture}, a high-level decision-making policy manages the \textit{target selection} subtask, while a low-level control policy governs the \textit{target driving} subtask.
This task decomposition aligns with existing methods in the literature on shepherding strategies for non-cohesive targets \cite{aulettaHerdingStochasticAutonomous2022, ninoDeepAdaptiveIndirect2023}. Each layer is trained in a specific herding scenario: the low-level policy is learned in a single-herder, single-target setting, while the high-level policy is trained in a multiple-herder, multiple-target environment with the previously learned low-level policy kept fixed. The resulting hierarchical control architecture enables each herder to independently select a target using the high-level policy and then interact with it effectively through the guidance of the specialized low-level policy.

In what follows, we leverage a continuous action space with PPO \cite{schulmanProximalPolicyOptimization2017} to learn both low-level and high-level policies, overcoming the limitations of previous approaches such as \cite{napolitano2024emergentcooperativestrategiesmultiagent}. We extend traditional PPO to cooperative multi-agent settings using Multi-Agent Proximal Policy Optimization (MAPPO) \cite{yuSurprisingEffectivenessPPO2022}. To rigorously validate our approach, we compare our trained policy against the state-of-the-art model-based strategy proposed in \cite{lamaShepherdingControlHerdability2024}.

\subsection{Metrics}
For validation we define $\chi(k)$ as the fraction of targets inside the buffered goal region at time step $k$:
\begin{equation}
    \chi(k) = \frac{ \mid \{ a : \mathbf{T}_a(k) \in  \hat{\Omega}_\R{G}, \; a \in \left[1, M\right]\} \mid}{M},
\end{equation}
where $\mid \mathcal{A} \mid$ denotes a set $\mathcal{A}$ cardinality.
The settling time $n^{\star}$ is the first time step when all targets enter and remain in $\hat{\Omega}_\R{G}$ :
\begin{equation}
    n^{\star} = \inf_{n} \{ n \geq 0  \; \text{:} \; \chi(k) \geq 0.99, \forall k \in \left[ n, n_\R{f} \right] \},
\end{equation}
where $n_\R{f} = \min \left( n + n_\R{t}, n_\R{h} \right)$. 
An episode terminates when all targets remain within the goal region for $n_\R{t}$ consecutive steps or if the maximum step count $n_\R{h}$ is reached.
The problem is solved if $n^\star$ is finite.

We assess efficiency of the policy using the path length, defined as the average distance traveled by the herders over a finite time interval $\left[0, n \right]$:
\begin{equation}
\label{eq:path_length}
    d(n) = \frac{1}{N} \sum_{i=1}^N \sum_{k=0}^{n-1} \norm{\mathbf{H}_i(k+1)-\mathbf{H}_i(k)}.
\end{equation}
In our simulations, we evaluate the path length $d = d(n_\R{f})$, which also indicates the scaled average control effort, given the herder dynamics in \eqref{eqn: herder_dyn}.

\section{Driving sub-task: one herder-one target}
\label{sec:driving}
The low-level policy for the driving subtask is trained in a single-herder, single-target scenario ($N=1, \ M=1$) with normalized target position and herder-target relative position as inputs.

While simple reward signals can theoretically produce complex behaviors, carefully designed reward functions enhance convergence and performance \cite{heessEmergenceLocomotionBehaviours2017}. Our reward function (inspired from the one used in \cite{albi2016}) balances four key objectives for the herder: (i) it should approach the target to enter its influence zone, (ii) position itself to guide the target toward the goal, (iii) minimize control effort, and (iv) recognize goal achievement. 
Hence, the reward function is shaped as:
\begin{equation}
    \begin{aligned}
    r_{\R{D}, k} =& - k_1 \norm{\B{T}(k) - \B{H}(k)} \indicator{\B{T}(k) \in \Omega_\R{G}} + \\ & - k_2 \left( \norm{\B{T}(k)} - \rho_G \right) \indicator{\B{T}(k) \in \Omega_\R{G}} -  k_3 \norm{\B{u}(k)}
    \end{aligned}
    \label{eqn:reward_1H1T}
\end{equation}
where $\indicator{x}$ is 1 when $x$ is true and 0 otherwise.

As reported in the Appendix, the gains $k_1$, $k_2$, $k_3$ were systematically determined by their relative importance: $k_2$ (goal guidance) received the highest value as our primary objective, $k_1$ (target approach) was set to an intermediate value to bootstrap early learning, and $k_3$ (efficiency) was assigned the lowest value as a secondary objective. This hierarchy ($k_3 < k_1 < k_2$) creates a natural progression that guides policy development through increasingly complex behaviors.

\begin{figure}[t]
    \centering
    \vspace{0.3cm}
    \subfloat[]
    {
        \includegraphics{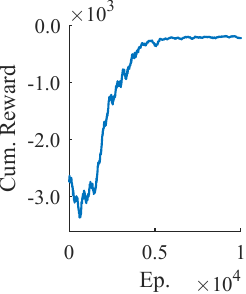}
        \label{fig:driving_rw}
    }
    \subfloat[]    
    {
        \includegraphics{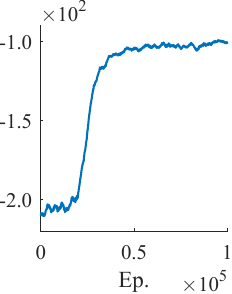}
        \label{fig:target_selection_rw}
    }
    \caption{Learning curves during training: \textbf{(a)} \textit{Driving} policy cumulative reward, smoothed via moving average of 200 samples; \textbf{(b)} \textit{Target-selection} policy cumulative reward, smoothed via moving average of 2000 samples. For both policies only the first half of the training is shown, to highlight the learning phase.}
    \label{fig:rewards_new}
\end{figure}

In our implementation of PPO, both the actor and the critic neural networks feature four input neurons and five hidden layers each made of 64 neurons with ReLU activation. As output, the critic network has a single neuron with a linear activation function, while the actor network has two neurons with tanh activation functions. During training, control actions are sampled from Gaussian distributions whose means are provided by the actor's output, while their standard deviations are extra parameters of the actor network subject to the optimizer updates \cite{schulmanTrustRegionPolicy2017}. Training occurred over $E=2 \cdot 10^4$ episodes of $n_\R{h}=1200$ steps each, with $n_\R{t} = 200$ steps. The training curve in Fig. \ref{fig:rewards_new}a shows rapid initial improvement followed by stable convergence, indicating effective reward shaping.

Our PPO implementation follows \cite{andrychowiczWhatMattersOnPolicy2020}, with hyperparameters initially based on \cite{schulmanProximalPolicyOptimization2017} and refined through systematic experimentation to optimize sample efficiency and learning stability.

\subsection{Validation}
\label{section:driving_validation}
The policy was tested across \(E=1000\) episodes and compared to the model-based driving policy proposed in \cite{lamaShepherdingControlHerdability2024},
where the herder steers the target by maintaining a fixed distance behind it, applying a proportional control strategy.

\begin{figure}
    \centering
        \vspace{0.2cm}
        \includegraphics{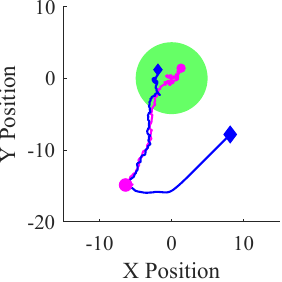}
    \caption{Example of the learned \textit{driving} policy in a single herder, single target setting: the herder (blue diamond) approaches the target (magenta circle), drives it to the goal region (green circle) and contains it. Big markers show initial positions, small ones show final positions.}
    \label{fig:driving_task_example}
\end{figure}

As illustrated in Fig.~\ref{fig:driving_task_example}, our RL-driven herder agent initially approaches the target, guides it toward the goal region, and then maintains its position using minimal movements, demonstrating the effectiveness of our reward formulation.

While both strategies successfully completed the task in all the episodes, Fig. \ref{fig:metrics_LL} shows how our RL policy significantly outperformed the heuristic strategy both in performance and efficiency.

\section{Multiple herders - Multiple targets Scenario}
\label{sec: target_selection}
In the case $N>1, M>1$, a MAPPO agent is trained for the target selection subtask. The learning process follows the centralized-training, decentralized-execution paradigm with parameter sharing \cite{guptaCooperativeMultiagentControl2017}. Specifically, during training, all herders contribute to learning a shared policy by updating the same neural network parameters. Then, once trained, each herder independently exploits this policy to select a target.
Due to constraints in the neural network structure on input and output dimensions, we consider the specific case of $N=2$ herders and $M=5$ targets.
Each herder receives as input its own position, the position of the other herder, and the positions of all targets in the environment, and then outputs the index of the selected target.
It is important to note that decision-making is decentralized, as each herder senses the positions of other agents, but selects a target independently and without communication with others.

For target selection, we define the reward function $r_{\R{T},k}$ as:
\begin{equation}
    \label{eqn:reward_L2}
    r_{\R{T}, k} = - k_4 \sum_{a=1}^M (\norm{\B{T}_a(k)} - \rho_\R{G}) \indicator{\B{T}_a(k) \in \Omega_\R{G}}
\end{equation}
Thus, at time \( k \), the agents receive an increasing penalty for each target located outside the goal region, scaled by $k_4 > 0$.

The MAPPO agent is implemented using an Actor-Critic neural network architecture. Both networks have $2(N+M)$ input neurons, representing the absolute positions of all the agents normalized with respect to the arena length, followed by $2$ hidden layers of $256$ and $128$ neurons, respectively, each activated by ReLU functions. The critic outputs the state value through a single neuron, while the actor has $M$ output neurons corresponding to the selectable targets. 
At each step, the outputs are normalized to sum to $1$, representing the probability distribution over target selection.

The target selection policy is trained over $E=2 \cdot 10^5$ episodes, each lasting $n_\R{h} = 3000$ steps with $n_\R{t}=200$ steps.

To facilitate training, we leverage the modularity of our hierarchical control architecture by fixing target selection for \(n_\R{w} = 100\) consecutive time steps, allowing sufficient time to influence targets and evaluate the selection's effectiveness. 
After training, we relax this assumption, allowing the herder to select a target at each time step.

In Fig. \ref{fig:rewards_new}b, we show the cumulative reward curve, while the corresponding reward weights and MAPPO hyperparameters are listed in the Appendix. Hyperparameters tuning largely began from the values used in the \textit{driving} policy, as these proved effective. Adjustments were made to the horizon and the number of actors, reflecting the reduced number of forward steps per episode. Additionaly, the entropy coefficient was set to zero, as the policy stochasticity itself provided sufficient exploration. Following \cite{yuSurprisingEffectivenessPPO2022}, the use of minibatches was discarded.

\subsection{Validation}
The learned policy is validated over \(E=1000\) episodes with seeded initial conditions uniformly sampled in $\Omega_0$. Each episode lasts $n_\R{h} = 3000$ steps and may terminate early if $n_\R{t} = 200$ successful steps are reached.

For comparison, we use the target selection law from \cite{lamaShepherdingControlHerdability2024}. In this approach, the $i$-th herder selects the target $\B{T}_i^*$ furthest from the goal region, satisfying  $\norm{\B{T}_i^* - \B{H}_i} \leq \norm{\B{T}_i^* - \B{H}_j}, \ \forall i \neq j$.
This rule partitions the set of selectable targets among herders, enforcing cooperation in target selection without requiring explicit communication, but only positional information. Each target-selection strategy is combined with the corresponding driving policy. 

RL agents can effectively learn decision-making strategies in a model-free fashion, as proven by Fig. \ref{fig:metrics_HL}: both strategies successfully solve all the episodes, but the RL agent achieves better settling times and path lengths.

\begin{figure}
    \centering
    \vspace{0.2cm}
        \includegraphics{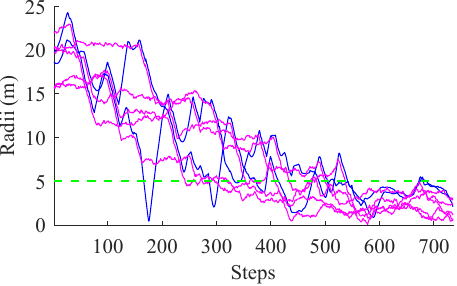}
    \caption{Validation example of the fully learning-based solution in the $N=2, M=5$ scenario: the radii of the herders (blue lines) and targets (magenta lines) are shown, compared to the goal region radius $\rho_\R{G} = 5$ (green dashed line). The herders successfully steer and contain the targets in the goal region.}
    \label{fig:target_selection_task_example}
\end{figure}

Fig. \ref{fig:target_selection_task_example} illustrates an example episode in which two herders successfully steer five passive agents toward the goal region and contain them, as indicated by the targets’ radii remaining within the goal region radius.

\subsection{Robustness analysis}
To assess the robustness of the trained policies under parametric variations, the target noise $D$, repulsion distance $\lambda$, and repulsion strength $k^\mathrm{T}$ are varied according to normal distributions, centered at the nominal values provided  in the Appendix, with a standard deviation of $30\%$. The evaluation is conducted over $E = 1000$ validation episodes with seeded initial conditions. 

\begin{figure}[t]
    \centering

    \vspace{0.5em}
    \subfloat[]
    {
        \includegraphics{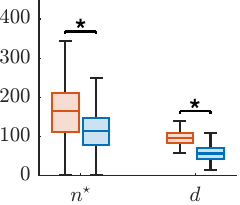}
        \label{fig:metrics_LL}
    }
    \subfloat[]
    {
        \includegraphics{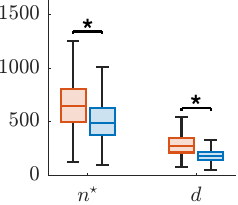}
        \label{fig:metrics_HL}
    }\\
    \subfloat[]
    {
        \includegraphics{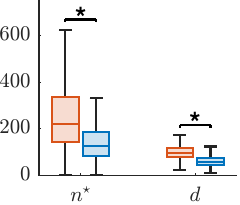}
        \label{fig:metrics_LL_ROB}
    }
    \subfloat[]
    {
        \includegraphics{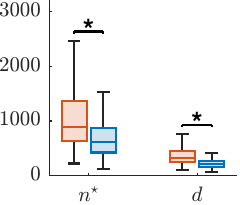}
        \label{fig:metrics_HL_ROB}
    }
    \caption{Validation results for the learning-based strategies (blue) over 1000 episodes with seeded initial conditions, showing average settling time $\smash{n^\star}$ in steps and average path length $d$ in meters. Compared against the heuristic approach (orange) from \cite{lamaShepherdingControlHerdability2024} for both the \textit{(a)}~$N=1,\ M=1$ and \textit{(b)}~$N=2,\ M=5$ configurations. A robustness analysis is also presented by varying the targets’ model parameters by 30\% around their nominal values, for both \textit{(c)}~$N=1,\ M=1$ and \textit{(d)}~$N=2,\ M=5$ settings. Box plots are shown for each metric. Mann-Whitney U test was performed on each metric pair yielding $p$-values always smaller than $0.001$.
}
    \label{fig:metrics}
\end{figure}

Numerical results for the scenarios with $N=1, M=1$ and $N=2, M=5$ are illustrated in Fig.~\ref{fig:metrics_LL_ROB} and Fig.~\ref{fig:metrics_HL_ROB}, respectively. In both scenarios, the RL-based approach demonstrates a significant improvement over the heuristic strategy, achieving superior performance in terms of both settling time and path length. Furthermore, the RL method consistently attains a success rate exceeding $97.50\%$, markedly higher than the heuristic approach's success rate of $84.20\%$. Comparable outcomes are also observed when considering a $20\%$ variation in the system parameters, where the RL strategy again shows greater robustness relative to the heuristic method.

\subsection{Extension to large-scale settings}
\label{sec:topological_sensing}
\begin{figure}[t]
    \centering
    \vspace{0.2cm}
    \includegraphics{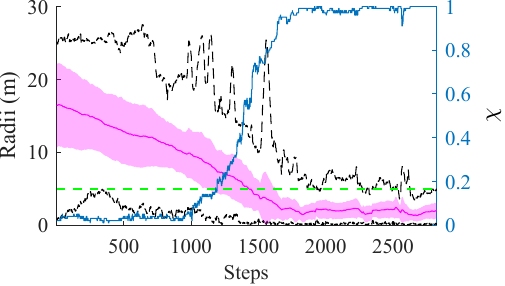}
    \caption{Example of the policy behavior in a $N=10, M=100$ setting, where the previously trained $\hat{N}=2, \hat{M}=5$ policy is extended using topological sensing for the herders. The figure depicts the evolution of the targets' radii, showing the mean (magenta line), standard deviation (magenta shading), minimum and maximum values (black dashed lines) and the fraction of captured targets $\chi$ (blue line), alongside the goal region radius $\rho_\R{G}$ (green dashed line).}
    \label{fig:topological_sensing}
\end{figure}

Unlimited sensing capabilities for herders is often unrealistic in practical applications. To address this limitation, we implement topological sensing (c.f. \cite{balleriniInteractionRuling2008}) where each herder perceives only the $\hat{N} \leq N$
nearest herders (including itself) and the $\hat{M} \leq M$ closest targets. This approach allows us to extend our target-selection policy trained for ($\hat N, \hat M$) agents to much larger environments ($N \gg \hat{N}, M \gg \hat{M}$), overcoming the neural network's fixed structure constraint. Fig. \ref{fig:topological_sensing} demonstrates this extension with $N=10$ herders managing $M=100$ targets using a policy originally trained for $\hat{N}=2, \hat{M}=5$ agents. Results show all targets ultimately reach the goal region, though some containment challenges exist due to limited sensing, similar to the herdability issues introduced and studied in \cite{lamaShepherdingControlHerdability2024}. Future work will focus on more robust, scalable approaches to address these remaining limitations.

\section{Conclusion}
In this work, we advance decentralized learning-based shepherding control by leveraging Proximal Policy Optimization to handle continuous action spaces. Each herder operates independently without explicit communication, effectively solving the shepherding task in a fully model-free manner. Our solution demonstrates scalability to larger scenarios while maintaining a restricted observation space.
The hierarchical architecture highlights the potential of advanced reinforcement learning in multi-agent control systems. However, the theoretical gap between our hierarchical solution and the optimal control strategy remains an open question. Future work will focus on: (1) extending to truly large-scale systems; (2) exploring heterogeneous agents with varying dynamics; (3) investigating adversarial target behaviors; (4) developing theoretical bounds on the optimality gap; and (5) validating with physical robots. Additionally, integrating limited communication protocols could potentially enhance coordination while maintaining computational efficiency.
\appendix
\renewcommand{\thetable}{A\arabic{table}}
\setcounter{table}{0} 
The parameters used for the numerical simulations described in Sections \ref{sec:driving}-\ref{sec: target_selection} and the hyperparameters of the PPO and MAPPO algorithms are reported in Tables \ref{tab:model_params} and \ref{tab:hyperparameters_ppo_mappo}, respectively. Vidoes are available at \href{https://github.com/SINCROgroup/Hierarchical-Policy-Gradient-RL-for-Multi-Agent-Shepherding-Control-of-Non-Cohesive-Targets}{https://github.com/SINCROgroup/Hierarchical-Policy-Gradient-RL-for-Multi-Agent-Shepherding-Control-of-Non-Cohesive-Targets}. 

This paper's English text has been revised with assistance from Claude \cite{claude2025}, an AI assistant developed by Anthropic, to improve clarity and readability.

\begin{table}
    \centering
    \vspace{0.1cm}
    \caption{Simulation parameters for the shepherding model derived from \cite{lamaShepherdingControlHerdability2024}. The first-order model is integrated via Euler-Maruyama method with a sampling time $\Delta t=0.05$ s.}
    \begin{tabular}{ll|ll}
        \toprule
        Parameter & Value &Parameter & Value \\
        \midrule
        $\rho_\R{G}$ & 5 & $D$ & 0.5 \\
        $R$ & 25 & $\lambda$ & 2.5 \\
        $v_\R{H}$ & 8 &$k^\R{T}$ & 3  \\
        
        \bottomrule
        
    \end{tabular}
    
    \label{tab:model_params}
\end{table}

\begin{table}
    \centering
    \caption{Hyperparameters and reward gains of the PPO/MAPPO training algorithms (values for MAPPO are indicated in parentheses only when different from the PPO ones). The parameters names refer to the nomenclature found in \cite{schulmanProximalPolicyOptimization2017}.}
    \begin{tabular}{ll|ll}
        \toprule
        Hyperparameter & Value & Gain & Value \\
        \midrule
        Adam stepsize & 5e-4 & $k_1$ & 5e-2\\
        Discount & 0.98 & $k_2$ & 1e-1\\
        GAE parameter & 0.95  & $k_3$ & 1.5e-2\\
        Clipping parameter & 0.2 & $k_4$ & 1e-2\\
        VF coeff. & 0.5 & &\\
        Entropy coeff. & 0.1 (0) & &\\
        Number of epochs & 10 & &\\
        Horizon & 4096 (32) & &\\
        Minibatch size  & 128 (1024) & &\\
        Number of actors & 8 (32) & & \\
        \bottomrule
        
    \end{tabular}
    
    \label{tab:hyperparameters_ppo_mappo}
\end{table}

\bibliographystyle{ieeetr}

\end{document}